\documentclass[conference]{IEEEtran}
\IEEEoverridecommandlockouts
\usepackage{cite}
\usepackage{amsmath,amssymb,amsfonts}
\usepackage{algorithmic}
\usepackage{graphicx}
\usepackage{textcomp}
\usepackage{xcolor}
\def\BibTeX{{\rm B\kern-.05em{\sc i\kern-.025em b}\kern-.08em
    T\kern-.1667em\lower.7ex\hbox{E}\kern-.125emX}}
\begin{document}

\title{Sustainable Face Recognition on Low-Power Devices with VQ-VAE Embeddings\\
}

\author{\IEEEauthorblockN{Christos Chronis}
\IEEEauthorblockA{\textit{Dep. of Informatics and Telematics} \\
\textit{Harokopio University of Athens}\\
Athens, Greece \\
chronis@hua.gr}
\and
\IEEEauthorblockN{Georgios Th. Papadopoulos}
\IEEEauthorblockA{\textit{Dep. of Informatics and Telematics} \\
\textit{Harokopio University of Athens}\\
Athens, Greece \\
g.th.papadopoulos@hua.gr}
\and
\IEEEauthorblockN{Iraklis Varlamis}
\IEEEauthorblockA{\textit{Dep. of Informatics and Telematics} \\
\textit{Harokopio University of Athens}\\
Athens, Greece \\
varlamis@hua.gr}
% \and
% \and
% \and
}

\maketitle

\begin{abstract}
Face recognition has become a cornerstone of modern AI applications, yet conventional approaches often rely on computationally intensive models deployed in cloud environments, leading to increased network traffic, high energy consumption, and a heavy carbon footprint. This work introduces a sustainable, edge-deployable face recognition framework based on Vector-Quantized Variational Autoencoders (VQ-VAE), which generates compact and semantically rich latent representations of facial images. By leveraging the compression capacity and reconstruction quality of VQ-VAE embeddings on the edge and combining them with the power of pre-trained face embeddings in a knowledge distillation setup, our system achieves comparable accuracy to state-of-the-art face embedding models while significantly reducing memory and computation requirements on the edge, making it suitable for low-power edge devices. 
The integration of VQ-VAE compression minimizes network overhead while keeping the matching accuracy high by retaining only the most informative facial features in the latent space. As a result, the reconstructed images preserve the key identity characteristics, improving the robustness and overall performance of the face embeddings.
\end{abstract}


\section{INTRODUCTION}

The increasing demand for secure and efficient access control in workplaces, residential buildings, and protected facilities has positioned face recognition systems (FRS) as one of the most promising biometric solutions. Unlike traditional access methods such as passwords or access cards, face recognition offers a contactless, non-intrusive, and highly accurate method of verifying individual identity. By leveraging modern AI techniques, such systems can automatically identify or verify a person, minimizing human errors and enhancing security. Deploying these systems on edge devices allows real-time inference without reliance on cloud infrastructure, which is critical for both latency-sensitive applications and the protection of sensitive biometric data.

Face recognition in practical scenarios faces significant challenges due to variations in lighting, posture, facial expressions, occlusions, and background conditions. An effective FRS must robustly handle these variations while maintaining high accuracy. Traditionally, achieving such performance requires large, computationally intensive models, such as FaceNet \cite{schroff2015facenet}, ArcFace \cite{deng2019arcface}, or SphereFace \cite{liu2017sphereface} that are deployed on powerful servers. However, this approach leads to high energy consumption and limits deployment in resource-constrained environments. By moving inference to the edge or leveraging compact models, it becomes possible to perform face recognition efficiently, reducing both computational cost and environmental impact while maintaining reliable identification.
We present an edge-deployable face recognition system built on Vector-Quantized Variational Autoencoders (VQ-VAE) \cite{van2017neural}, which compress facial images into compact, semantically meaningful latent codes. This allows low-power devices to transmit only compressed representations while preserving identity-relevant information.

Images captured on the edge are first processed for face detection, and only the detected regions are encoded by the VQ-VAE. The resulting discrete codes are sent to the cloud, where the decoder reconstructs the faces and a recognition model extracts embeddings for identity matching. The VQ-VAE is fine-tuned via knowledge distillation to improve reconstruction quality and retain discriminative details needed for recognition.

\begin{figure}[!t]
    \centering
    \includegraphics[width=\linewidth]{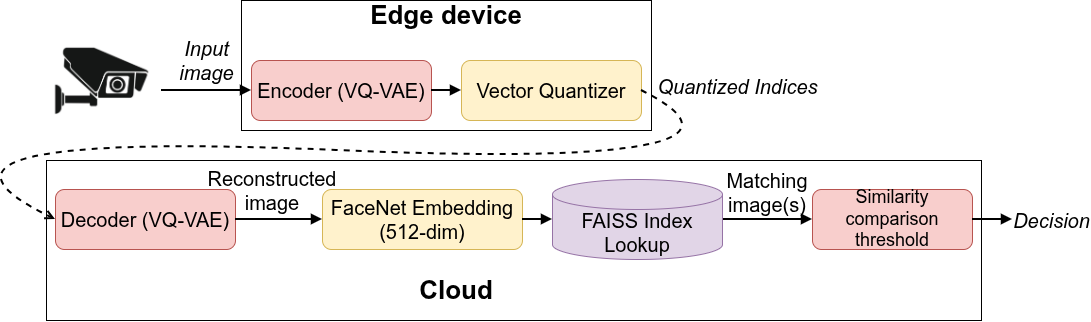}
    \caption{Workflow of the proposed Face Recognition approach. The system performs local face detection and encoding, sends the VQ-VAE quantized indices to the cloud, where the image is reconstructed. The reconstructed image embeddings are then used to retrieve the best candidate image in a vector database and decide upon the retrieved matches similarity.}
    \label{fig:edge_vqvae_workflow}
\end{figure}

By shifting compression to the edge and reconstruction and embedding generation to the cloud, the system reduces on-device computation and memory use, conserves bandwidth, and maintains high recognition accuracy. Figure \ref{fig:edge_vqvae_workflow} summarizes the workflow and its integration of efficient models, semantic indexing, and on-device decision-making.

Compared to conventional pretrained embedding models, the VQ-VAE approach achieves high recognition accuracy while greatly reducing edge-side memory and computation. Knowledge distillation from a larger model further strengthens the VQ-VAE, ensuring its compressed latent codes retain the most discriminative facial features. By performing compression locally, the system minimizes data sent to the cloud, lowering network overhead while preserving the information required for reliable reconstruction and embedding generation.

This design supports Sustainable and Trustworthy AI by optimizing resource use and reducing environmental impact. The division of labor between edge and cloud decreases energy consumption on the device while maintaining strong identification performance. Distilled VQ-VAE models ensure identity-relevant information is preserved with minimal transmission, enabling secure and efficient face matching. Overall, the framework demonstrates how modern AI techniques can deliver responsible, sustainable, and effective access-control solutions.

The remainder of this paper is organized as follows. Section \ref{sec:rel_work} reviews related work in face recognition systems, edge-based inference, and sustainable AI approaches. Section \ref{sec:approach} presents the proposed approach, detailing the system architecture, VQ-VAE-based embedding generation, semantic indexing, and on-device access control workflow. Section \ref{sec:experiments} describes the experimental evaluation, including the dataset used to train the VQ-VAE, the test samples for performance and efficiency assessment, and comparisons with alternative methods based on pretrained image embeddings. Section \ref{sec:results} reports the experimental results and discussion, highlighting the advantages of the proposed VQ-VAE approach in terms of accuracy, resource usage, and sustainability, as well as its performance relative to conventional pretrained embedding models. Finally, Section \ref{sec:conclusions} concludes the paper and highlights our future steps in this field.

\section{RELATED WORK}
\label{sec:rel_work}

The development of modern face recognition systems has been fundamentally shaped by the advent of deep learning. Foundational works like FaceNet \cite{schroff2015facenet} established the paradigm of learning a unified embedding space for faces using deep convolutional networks and triplet loss, while subsequent innovations in objective functions, such as the additive angular margin loss in ArcFace \cite{deng2019arcface} and the angular margin in SphereFace \cite{liu2017sphereface}, significantly enhanced feature discriminability. These approaches, often built on large-scale datasets like VGGFace \cite{parkhi2015deep} or VGGFace2 \cite{cao2018vggface2}, set a high bar for accuracy but typically rely on computationally intensive models deployed on powerful servers. This creates a tension between performance and practicality, underscoring the need for approaches that deliver high recognition accuracy while drastically reducing computational load, memory usage, and data transmission requirements.

To address the limitations of server-based models, significant effort has been dedicated to creating efficient and lightweight face recognition architectures. Works such as VarGFaceNet \cite{yan2019vargfacenet} and ShuffleFaceNet \cite{martindez2019shufflefacenet} exemplify this trend, employing techniques like variable group convolution and channel shuffling to reduce computational cost and model size while maintaining high accuracy. 
Similarly, MobileFaceNet \cite{sggan2019} has been proposed as an adaptation of SphereFace for face recognition, offering a compact and efficient architecture suitable for edge deployment.
This push for efficiency is intrinsically linked to the broader movement of on-device AI inference. The field of TinyML \cite{schizas2022tinyml} focuses on deploying neural networks on ultra-low-power microcontrollers, while surveys on on-device AI \cite{wang2025empowering} and efficient deep learning for edge inference \cite{murshed2021machine} outline the overarching challenges and solutions, including model compression and hardware-software co-design. Deploying FRS on the edge mitigates latency and privacy concerns by processing data locally, eliminating the need to transmit sensitive biometric information over networks.
Although these lightweight models offer substantial reductions in computation and memory, they generally experience a notable drop in performance compared to their full-scale counterparts.

However, the practical deployment of FRS also raises significant challenges related to trustworthiness and security. The vulnerability of deep learning models to adversarial attacks is a critical concern \cite{kilany2025comprehensive}, necessitating robust defense mechanisms. Furthermore, the push for trustworthy face recognition pipelines \cite{leong2025bridging} emphasizes the need for systems that are not only accurate and robust but also explainable and privacy-preserving. These trustworthiness concerns are complemented by the growing need for Sustainable AI \cite{wu2022sustainable}. The substantial energy consumption of large-scale AI models has prompted a critical examination of their environmental impact, with research indicating that future progress hinges on more efficient training and inference paradigms \cite{xiang2024enabling}.

Our work sits at the intersection of these research thrusts. We build upon the foundational principles of deep face recognition but move away from fully edge-based or purely server-centric pipelines. Instead, we leverage the Vector-Quantized Variational Autoencoder (VQ-VAE) \cite{van2017neural} to perform lightweight, on-device compression of facial images before they are transmitted to the cloud for reconstruction and embedding extraction. While our system still relies on a large recognition model in the cloud, we substantially reduce its environmental and privacy impact by running only lightweight VQ-VAE compression on the edge, transmitting minimal data, and reconstructing high-quality images that preserve accuracy without requiring full-resolution inputs. This hybrid strategy provides a practical pathway that balances the benefits of state-of-the-art recognition models with significantly reduced data transfer, lower on-device computation, and improved privacy, offering a realistic and impactful step toward sustainable and trustworthy face recognition. By designing this edge–cloud workflow for access control, we directly contribute to the fields of efficient edge inference \cite{zhang2025breaking} and trustworthy AI \cite{wang2024survey}. Finally, by prioritizing local processing, privacy-preserving data transfer, and computational efficiency, our system inherently supports the principles of Sustainable AI, demonstrating that high-performance face recognition can be achieved within environmentally conscious and user-centric design constraints.

\section{PROPOSED APPROACH}
\label{sec:approach}

The proposed system leverages a \textbf{Vector-Quantized Variational Autoencoder (VQ-VAE)} to enable efficient face recognition in edge–cloud deployments. Unlike conventional VAEs, which encode inputs into continuous latent distributions, VQ-VAE maps features to a discrete codebook, producing compact, semantically meaningful embeddings suitable for transmission and storage on low-power edge devices. This discrete latent space reduces redundancy and ensures that only the most informative features are preserved, facilitating efficient downstream processing.

To further enhance the quality of compressed representations, the VQ-VAE is trained with knowledge distillation from a larger, high-capacity model. This ensures that the discrete embeddings retain the most discriminative information necessary for reliable face recognition. On the edge, incoming images are encoded into these compact latent codes, which are then transmitted to the cloud. There, the VQ-VAE decoder reconstructs high-fidelity facial images, enabling pretrained recognition models such as FaceNet to extract embeddings that are accurate and comparable to those obtained from original images. 

By combining distillation-enhanced compression on the edge with reconstruction and embedding extraction in the cloud, the system achieves a practical balance between computational efficiency, bandwidth reduction, and recognition accuracy. The resulting workflow preserves identity-relevant features while minimizing network overhead and on-device computation, providing a sustainable and trustworthy foundation for secure, real-time face recognition.

\subsection{VQ-VAE Representation Learning}

Given an input face image $\mathbf{x} \in \mathbb{R}^{H \times W \times C}$, the encoder $E_{\theta}$ maps it into a continuous latent space $\mathbf{z}_e = E_{\theta}(\mathbf{x})$. Instead of directly using $\mathbf{z}_e$, VQ-VAE employs a discrete codebook $\mathcal{C} = \{\mathbf{e}_k\}_{k=1}^{K}$ with $K$ embedding vectors, each of dimension $D$. The encoder output is quantized by finding the nearest codebook entry according to the Euclidean distance:

\begin{equation}
\mathbf{z}_q = \mathbf{e}_{k^*}, \quad \text{where} \quad k^* = \arg\min_{k} \|\mathbf{z}_e - \mathbf{e}_k\|_2.
\end{equation}

The decoder $D_{\phi}$ reconstructs the original image from the quantized latent representation:

\begin{equation}
\hat{\mathbf{x}} = D_{\phi}(\mathbf{z}_q).
\end{equation}

The model is trained once in the cloud by minimizing a composite loss function:
\begin{equation}
\mathcal{L} = \|\mathbf{x} - \hat{\mathbf{x}}\|_2^2 + \|\text{sg}[\mathbf{z}_e] - \mathbf{e}_{k^*}\|_2^2 + \beta \|\mathbf{z}_e - \text{sg}[\mathbf{e}_{k^*}]\|_2^2,
\end{equation}
where $\text{sg}[\cdot]$ denotes the stop-gradient operator, and $\beta$ controls the commitment cost encouraging the encoder output to remain close to the chosen codebook vector.

After training, only the \textbf{encoder} $E_{\theta}$ and the learned \textbf{codebook} $\mathcal{C}$ are deployed on the edge device for inference, while the decoder $D_{\phi}$ is retained for potential reconstruction or debugging purposes in the cloud.
The total storage required on the edge is therefore given by $\mathcal{O}(|\theta| + K \cdot D)$, where $|\theta|$ denotes the number of encoder parameters and $K \cdot D$ corresponds to a codebook of $K$ embedding vectors, each of dimension $D$. Since typically $K \ll N$ (number of training samples) and $D$ is small, the overall memory footprint remains minimal—on the order of a few megabytes—making it feasible for deployment on low-power edge devices.

\subsection{Knowledge Distillation for Reconstruction Quality}

\begin{figure}[!t]
    \centering
    \includegraphics[width=\linewidth]{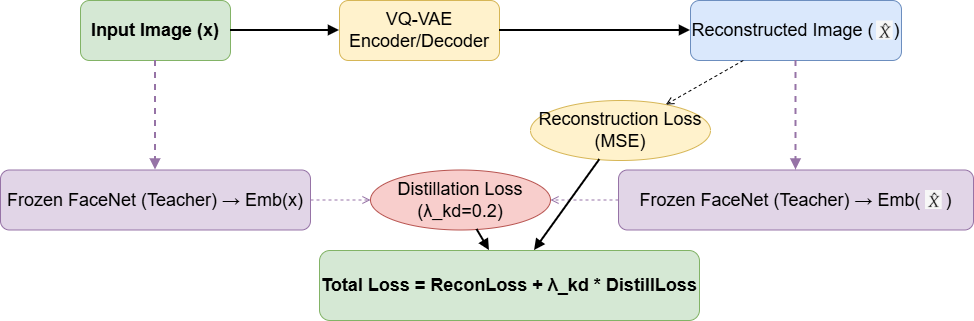}
    \caption{The process of VQ-VAE training using knowledge distillation from the FaceNet model.}
    \label{fig:knowledge_dist}
\end{figure}

To enhance the quality of VQ-VAE reconstructions and preserve identity-relevant features, we apply knowledge distillation from a pretrained face recognition model, denoted here as $F_{\psi}$ (e.g., FaceNet). An overview of the process is depicted in Figure \ref{fig:knowledge_dist}.

Given a training image $\mathbf{x}$, the VQ-VAE encoder and decoder produce a reconstruction $\hat{\mathbf{x}} = D_{\phi}(E_{\theta}(\mathbf{x}))$. To ensure that $\hat{\mathbf{x}}$ retains the facial semantics important for recognition, we compute embeddings from the pretrained model:

\begin{equation}
\mathbf{f}\text{teacher} = F{\psi}(\mathbf{x}), \quad
\mathbf{f}\text{student} = F{\psi}(\hat{\mathbf{x}}).
\end{equation}

The VQ-VAE is then trained to minimize a distillation loss that encourages the reconstructed image to produce embeddings close to those of the original image:

\begin{equation}
\mathcal{L}\text{KD} = |\mathbf{f}\text{teacher} - \mathbf{f}_\text{student}|_2^2.
\end{equation}

This distillation loss is combined with the standard VQ-VAE reconstruction loss:
% and commitment losses:\todo{this must match the figure notation}

\begin{equation}
\mathcal{L}\text{total} = 
% \mathcal{L}\text{Recon} + 
\mathcal{L}\text{VQ-VAE} + \lambda_\text{KD} \mathcal{L}_\text{KD},
\end{equation}

\noindent where $\lambda_\text{KD}$ balances the contribution of the knowledge distillation term. By training with this additional objective, the VQ-VAE learns to compress and reconstruct faces in a way that preserves identity-critical features, improving the semantic fidelity of the embeddings extracted from $\hat{\mathbf{x}}$ by any downstream face recognition model.

After training, the encoder $E_{\theta}$ and codebook $\mathcal{C}$ are deployed on the edge, while the decoder $D_{\phi}$ is retained in the cloud for reconstruction, ensuring that edge compression maintains the essential information needed for accurate cloud-based embedding extraction and matching.

\subsection{Image reconstruction and Semantic Matching}

During operation, a captured image $\mathbf{x}_{t}$ is processed locally on the edge device:
\begin{enumerate}
    \item The image undergoes face detection and normalization, where the system first detects the facial region using a lightweight detector (e.g., MTCNN or MobileFace) and then aligns and resizes it to a canonical format (fixed scale, rotation, and illumination) to ensure consistent embedding generation.
    \item The normalized face is passed through the encoder $E_{\theta}$, which transforms it into a discrete latent embedding $\mathbf{z}_q = E_{\theta}(\mathbf{x}_{t})$ using the quantized codebook representation.
     \item The resulting embedding $\mathbf{z}_q$ is transmitted to the cloud, where the VQ-VAE decoder reconstructs the facial image $\hat{\mathbf{x}}_t$ and a pretrained recognition model (e.g., FaceNet) extracts a high-quality embedding $\mathbf{f}t = F{\psi}(\hat{\mathbf{x}}_t)$ for semantic matching.

\end{enumerate}

The extracted embedding $\mathbf{f}_t$ is then compared against the set of stored embeddings ${\mathbf{f}i}{i=1}^{N}$ representing authorized individuals in the database. Semantic matching is performed using cosine similarity or Euclidean distance:

\begin{equation}
s(\mathbf{f}_t, \mathbf{f}_i) = \frac{\mathbf{f}_t \cdot \mathbf{f}_i}{|\mathbf{f}_t||\mathbf{f}_i|}, \quad i = 1, \ldots, N,
\end{equation}

and the identity $\hat{i}$ is determined by:

\begin{equation}
\hat{i} = \arg\max_{i} s(\mathbf{f}_t, \mathbf{f}_i).
\end{equation}

To further enhance robustness, a small subset of top-$k$ most similar embeddings can be retrieved. The corresponding reconstructed facial images can then be compared more precisely with the query image using fine-grained similarity measures in either the pixel or latent domain, and the consistency of semantic labels and confidence scores is used to validate the target identity.

In scenarios with highly constrained edge resources, only the VQ-VAE encoding step is performed locally, while reconstruction and FaceNet embedding extraction are entirely offloaded to the cloud. Importantly, both configurations preserve privacy and sustainability, as no raw facial images are transmitted — only compact latent codes are exchanged, substantially reducing energy consumption and network bandwidth.

\subsection{Advantages of VQ-VAE Embeddings on the Edge}

Conventional face recognition systems such as \textbf{FaceNet} or \textbf{ArcFace} rely on deep convolutional neural networks trained to map face images into high-dimensional continuous embeddings. During inference, a captured face is encoded into a vector in this embedding space, which is then compared against a database of known faces using distance metrics such as cosine similarity or Euclidean distance. These models typically require substantial computational resources and memory, and are often deployed on cloud servers to handle both encoding and similarity search. While highly accurate, cloud-based embeddings pose challenges in terms of latency, bandwidth, energy consumption, and privacy, as raw images or embeddings must be transmitted off-device for processing.

The VQ-VAE architecture, enhanced with knowledge distillation from a pretrained recognition model, provides several advantages over conventional pretrained embeddings:
\begin{itemize}
    \item \textbf{Compactness and Efficiency:} The discrete latent representation $\mathbf{z}_q$ is low-dimensional and efficiently encoded, reducing both computation and memory requirements on the edge.
    \item \textbf{Distillation-enhanced Semantic Fidelity:} By distilling identity-relevant features from a high-capacity model (e.g., FaceNet) during training, the reconstructed images retain the facial characteristics necessary for accurate cloud-side embedding extraction.
    \item \textbf{Sustainability:} Edge-based encoding minimizes data transmission, while the cloud reconstruction allows use of high-performance models without sending raw images, reducing energy consumption and network load.
    \item \textbf{Trustworthiness:} Since raw facial images are never transmitted, and only compact codes are exchanged, the system preserves privacy and aligns with ethical AI principles.
\end{itemize}

This hybrid edge–cloud workflow, combined with the training-once, deploy-many paradigm, ensures that the VQ-VAE encoder and codebook can be used across distributed devices efficiently, while the cloud-side reconstruction and FaceNet inference guarantee high recognition accuracy. Overall, this design aligns with the objectives of \textbf{Sustainable and Trustworthy AI}, providing a practical framework for energy-efficient, privacy-preserving, and high-performance face recognition in real-world access control applications.

\section{EXPERIMENTAL EVALUATION}
\label{sec:experiments}

The experimental evaluation aims to demonstrate the effectiveness and efficiency of the proposed edge-cloud face recognition system, utilizing VQ-VAE, knowledge distillation from a large model and face embeddings, and compare it with cloud-only or edge-only based alternatives. A well-curated dataset is essential for both training and evaluation. The dataset must fulfill three requirements: (i) sufficient images for training the VQ-VAE encoder to learn robust, semantically meaningful embeddings, (ii) images corresponding to specific "authorized user" IDs to populate the semantic database, and (iii) images corresponding to "unauthorized user" IDs used only for validating the system's ability to reject invalid users. This setup ensures that both identification and verification tasks can be rigorously evaluated while mimicking real-world access control scenarios.

\subsection{Training and test dataset}

For this study, we use the \textbf{CelebA} dataset~\cite{liu2015faceattributes}, a large-scale face dataset containing over 200,000 celebrity images with annotated identities and attributes. The dataset is split into a training set for training the VQ-VAE encoder and a held-out set for evaluation. A subset of identities from the held-out set is designated as \emph{authorized users}, whose embeddings are partially stored in the semantic database and partially used for testing, while the remaining identities are treated as \emph{unauthorized users} for testing to evaluate the system's rejection capability. All images are preprocessed with alignment, cropping, and resizing to match the input requirements of the VQ-VAE and alternative embedding models.

\begin{table}[t]
\centering
\caption{CelebA dataset splits for training and evaluation.}
\label{tab:celeba_stats}
\resizebox{\linewidth}{!}{
\begin{tabular}{lccc}
\hline
\textbf{Subset} & \textbf{\#Users} & \textbf{\#Images} & \textbf{\#Usage}   \\
\hline
VQ-VAE Train & 9,677 & 150,328 & Train encoder/decoder  \\
Authorized & 250 & 4,481  & Reference Images \\
&  & / 900 & / Test acceptance\\\\
Unauthorized & 250 &  900  & Test rejection \\
% Unauthorized & 250 & 4,317 train / 900 test  \\
\hline
Evaluation& 10,177 & 156,609 & ---  \\

\hline
\end{tabular}
}
\end{table}

\subsection{Model Architectures and Parameter Configuration}

To benchmark the proposed system, we evaluate three neural network architectures: VQ-VAE, FaceNet, and MobileFaceNet. The VQ-VAE is used on the edge side for compression and discrete encoding of face images of size $128 \times 128$ pixels, employing three input channels, a latent dimensionality of $D = 128$, and a codebook of $N = 256$ embeddings. The reconstructed images $x_{\text{recon}}$ are subsequently processed by the embedding networks for identity verification.

FaceNet, based on the InceptionResnetV1 architecture pretrained on the VGGFace2 dataset, operates on $160 \times 160$ pixel inputs and generates 512-dimensional discriminative embeddings. MobileFaceNet, used as an efficient lightweight baseline\footnote{https://github.com/cunjian/face\_rec\_MobileFaceNet}
, receives inputs of $112 \times 112$ pixels and produces 128-dimensional embeddings through a compact design employing InvertedResidual and GDConv blocks. Although FaceNet provides higher accuracy, its computational footprint limits deployment on edge devices, whereas MobileFaceNet offers a favorable trade-off between performance and efficiency.

The embeddings from both models are fine-tuned on a held-out subset of CelebA to classify face images as \emph{valid} or \emph{invalid}. During evaluation, embeddings for authorized users are stored in a semantic database, and query images are compared against this database for identity prediction, enabling adaptation of pretrained models to the distribution and requirements of the proposed verification task.

\subsection{Performance Metrics}

The performance evaluation of the proposed identity verification pipeline includes metrics that assess model training quality, verification accuracy, and computational efficiency for edge deployment.

\textbf{Training performance metrics:} The convergence of the VQ-VAE model is assessed through several indicators of reconstruction quality and representation stability. These include the Peak Signal-to-Noise Ratio (PSNR), the Structural Similarity Index Measure (SSIM), the code usage ratio (i.e., the proportion of active codebook entries out of $N = 256$), and the cosine similarity between the FaceNet embeddings of the original and reconstructed images, $\cos(f(x), f(x_{\text{recon}}))$. The latter reflects how well the reconstruction preserves identity-relevant features.

% \todo[inline]{a paragraph on the theoretical computation of network overhead}

\textbf{Verification performance metrics:} Identity verification accuracy is measured using a FAISS-based similarity search protocol on a deterministic test set containing 900 authorized and 900 unauthorized requests per split, where each request corresponds to an image embedding. A user image is labeled as authorized if its similarity with a retrieved image from the database exceeds a threshold of 0.65, a value determined experimentally by analyzing similarity distributions between genuine (same-identity) and impostor (different-identity) image pairs. The performance is measured using Accuracy, Recall, Precision and F-measure, under two user verification strategies: Top-1 and Top-5, which accept a user if the similarity with the most similar image in the database is above threshold, or when at least three of the top five similarity scores are above the threshold, respectively.

\textbf{Computational efficiency metrics:} To assess feasibility on edge hardware, efficiency measurements were collected over 100 inference runs (excluding warm-up) for the VQ-VAE, FaceNet, and MobileFaceNet models. Reported metrics include \textbf{model size} (MB), \textbf{peak memory usage} (MB), and average \textbf{inference time} (ms), quantifying the overhead of compression, reconstruction, and embedding extraction during deployment.

Together, these metrics provide a holistic evaluation of both training behavior and real-world operational performance, demonstrating the practicality of low-bandwidth identity verification within an edge–cloud architecture.

\section{RESULTS}
\label{sec:results}
\begin{figure}[!t]
    \centering
    \includegraphics[width=\linewidth]{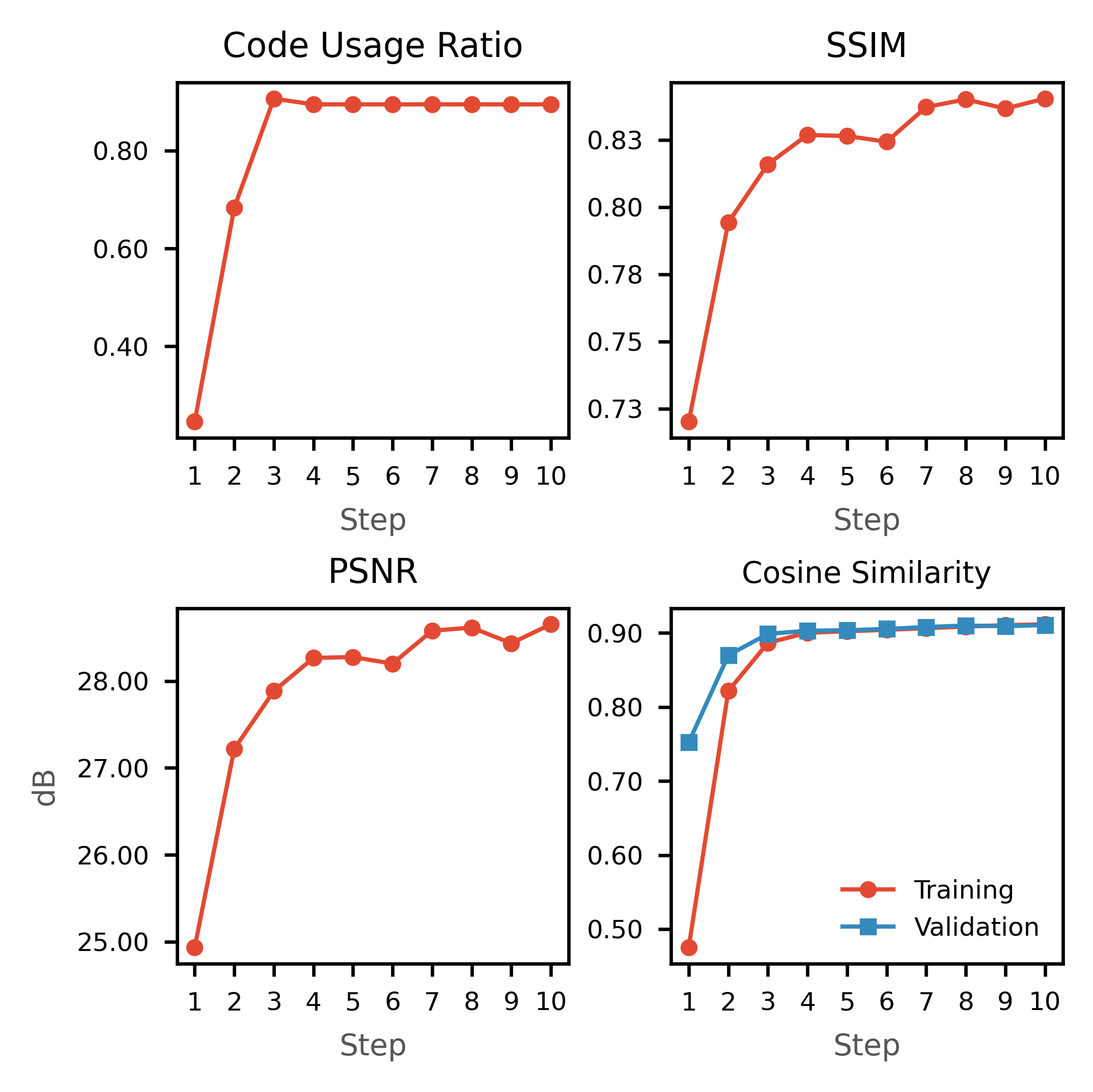}
    \caption{Training performance metrics. Code Usage Ratio, SSIM, and PSNR (dB) quantify reconstruction fidelity, while the Training vs. Validation cosine similarity measures feature alignment stability. The metrics demonstrate consistent performance improvements across steps.}
    \label{fig:training_metrics}
\end{figure}

Figure~\ref{fig:training_metrics} summarizes the convergence behavior of the VQ-VAE model over 10 optimization steps, evaluated using reconstruction quality and identity-preservation metrics. The \textbf{code usage ratio} increases sharply from approximately 0.27 in the first step to above 0.85 by step 3, indicating rapid stabilization of the discrete latent space and effective utilization of the codebook entries. Reconstruction fidelity also improves consistently, with \textbf{SSIM} rising from 0.73 to 0.84 and \textbf{PSNR} increasing from 25~dB to nearly 29~dB. To assess preservation of semantic identity relevant to face verification, embeddings of original and reconstructed images were extracted using FaceNet, and their \textbf{cosine similarity} was computed. This metric increases steeply during early training and stabilizes above 0.92 for both training and validation splits, demonstrating that the reconstructed images retain identity-discriminative information after compression and decoding. Collectively, the results confirm the rapid convergence of the VQ-VAE and its suitability as an edge-side encoder that preserves identity characteristics essential for downstream verification.

The memory requirements of the evaluated models are summarized in Table~\ref{tab:memory_requirements}. FaceNet exhibits the highest parameter footprint at 106.58 MB, reflecting its deep architectural complexity and its suitability for cloud-based identity embedding computation. MobileFaceNet provides a significantly smaller model size of 3.89 MB while maintaining competitive embedding capability, which makes it appropriate for embedded and low-power environments. The Depthwise VQ-VAE configuration, which performs edge-side compression and reconstruction, achieves exceptional compactness with a parameter size of only 0.23 MB, representing a reduction of more than two orders of magnitude relative to FaceNet. Its peak runtime memory requirement of 276.63 MB remains within acceptable limits for deployment on lightweight edge hardware. These findings demonstrate the value of depthwise-separable architectures combined with vector quantization for achieving identity-preserving compression under strict hardware constraints.

\begin{table}[h!]
\centering
\caption{Memory requirements}
\label{tab:memory_requirements}
\resizebox{0.7\linewidth}{!}{
\begin{tabular}{lcc}
\hline
\textbf{Model} & \textbf{Params (MB)} & \textbf{Peak (MB)} \\
\hline
FaceNet & 106.58 & 244.66 \\
MobileFaceNet & 3.89 & 250.35 \\
VQ-VAE & 0.23 & 276.63 \\
\hline
\end{tabular}}
\end{table}

Inference time performance results are presented in Table~\ref{tab:inference_time}, covering evaluation on CPU, GPU, and Raspberry Pi 4 platforms. All values are means within the 95\% CI. MobileFaceNet achieves the fastest execution across all hardware configurations, with mean times of 3.26 ms on GPU and 7.78 ms on CPU, indicating strong suitability for real-time authentication on embedded systems. The VQ-VAE also performs efficiently, requiring only 0.71 ms on GPU and 5.08 ms on CPU, and maintains low latency even on the Raspberry Pi 4 at 8.38 ms. In contrast, FaceNet incurs substantially higher inference cost, reaching 27.69 ms on CPU and 70.47 ms on Raspberry Pi 4. These results indicate that full-scale face embedding extraction with FaceNet is more appropriate for cloud-based processing, while the combination of Depthwise VQ-VAE and MobileFaceNet supports practical deployment for edge-side encoding and preliminary identity screening.

\begin{table}[h!]
\centering
\caption{Inference time evaluation}
\label{tab:inference_time}
\resizebox{0.9\linewidth}{!}{
\begin{tabular}{lcccccc}
\hline
&
\textbf{FaceNet} &
\textbf{MobileFaceNet} &
\textbf{VQ-VAE} \\
\textbf{Device} & \textbf{Mean (ms)} & 
  \textbf{Mean (ms)} & 
  \textbf{Mean (ms)}  \\
\hline
CPU  & 27.69 $\pm$ 0.202 & 7.78 $\pm$ 0.050 & 5.08 $\pm$ 0.095 \\
GPU  & 9.83  $\pm$ 0.133 & 3.26 $\pm$ 0.022 & 0.71 $\pm$ 0.008 \\
Raspberry Pi 4 & 70.47 $\pm$ 0.226 & 11.99 $\pm$ 0.010 & 8.38 $\pm$ 0.089 \\
\hline
\end{tabular}}
\end{table}

To estimate the network overhead of the proposed system, we consider the amount of data that must be transmitted for identity verification. For the VQ-VAE pipeline, only the discrete latent codes of dimension $D = 128$ with a codebook of $N = 256$ entries are sent from the edge to the cloud. Representing each code index with 1 byte, a $128 \times 128$ RGB image would be compressed into a sequence of 128 discrete indices, resulting in a total transmission of approximately 128 bytes per face image, plus negligible overhead for metadata. In contrast, full-resolution images required by FaceNet ($160 \times 160 \times 3$ pixels, 8-bit per channel) would require 76,800 bytes per image if transmitted directly to the cloud. MobileFaceNet ($112 \times 112 \times 3$ pixels) reduces this to 37,632 bytes per image. Consequently, the VQ-VAE approach theoretically reduces network transmission by over two orders of magnitude compared to raw images for FaceNet and by nearly three orders of magnitude compared to typical cloud-based image input, while still preserving the identity information necessary for accurate embedding extraction. This demonstrates that edge-side compression using discrete latent representations can drastically minimize network load, enabling scalable deployment in bandwidth-constrained or privacy-sensitive environments.

Table~\ref{tab:top1-summary} and Table~\ref{tab:top5-summary} present the average performance of the proposed VQ-VAE model compared to FaceNet and MobileFaceNet in five evaluation runs. All results are reported at 95\% CI. While FaceNet achieves the highest accuracy, our proposed approach reaches 0.72 Top-1 and 0.73 Top-5 accuracy, providing a strong performance level considering its lightweight nature. Unlike traditional models that require transmitting full feature representations, the VQ-VAE model generates compact quantized indices, enabling efficient edge-to-cloud processing with significantly reduced bandwidth requirements. In contrast, MobileFaceNet, despite being optimized for mobile hardware, demonstrates poor accuracy and unstable performance. These results show that the proposed method offers an effective balance between recognition accuracy and communication efficiency, making it highly suitable for real-world edge-based facial recognition systems.

\begin{table}[h!]
\centering
\caption{Verification Performance Top-1}
\resizebox{\linewidth}{!}{
\begin{tabular}{lcccc}
\hline
\textbf{Model} & \textbf{Accuracy} & \textbf{Precision} & \textbf{Recall} & \textbf{F1} \\
\hline
VQ-VAE  & $0.72 \pm 0.008$ & $0.67 \pm 0.013$ & $0.85 \pm 0.020$ & 0.75 \\
FaceNet          & $0.81 \pm 0.008$ & $0.75 \pm 0.007$ & $0.91 \pm 0.011$ & 0.82\\
MobileFaceNet    & $0.50 \pm 0.000$ & $0.50 \pm 0.000$ & $1.00 \pm 0.000$ & 0.67 \\

\hline
\end{tabular}}
\label{tab:top1-summary}
\end{table}

\begin{table}[h!]
\centering
\caption{Verification Performance Top-5}
\resizebox{\linewidth}{!}{
\begin{tabular}{lcccc}
\hline
\textbf{Model} & \textbf{Accuracy} & \textbf{Precision} & \textbf{Recall} & \textbf{F1} \\
\hline
VQ-VAE  & $0.73 \pm 0.005$ & $0.79 \pm 0.021$ & $0.63 \pm 0.023$ & 0.71 \\
FaceNet          & $0.84 \pm 0.011$ & $0.89 \pm 0.010$ & $0.78 \pm 0.013$ & 0.83\\
MobileFaceNet    & $0.50 \pm 0.000$ & $0.50 \pm 0.000$ & $1.00 \pm 0.000$ & 0.67 \\

\hline
\end{tabular}
\label{tab:top5-summary}}
\end{table}

\section{CONCLUSIONS}
\label{sec:conclusions}

This work presented a hybrid face-recognition pipeline that leverages a compact VQ-VAE encoder on the edge and a cloud-based teacher model to achieve efficient, privacy-preserving, and identity-consistent verification. The results demonstrate that the proposed approach offers strong reconstruction quality, high identity preservation, and fast inference, while requiring only a fraction of the memory footprint of conventional embedding models. The VQ-VAE’s lightweight architecture, combined with discrete semantic representations, enables real-time operation on embedded hardware and significantly reduces both energy consumption and data transmission requirements.

By integrating edge compression with teacher-guided reconstruction, the system provides a practical pathway toward sustainable and trustworthy face recognition without compromising verification performance. Future work will extend this framework by exploring adaptive codebook learning, improved reconstruction fidelity, and tighter integration with security-critical components such as spoof detection and confidence modeling, further advancing the deployment of responsible AI in resource-constrained environments.

\section*{Acknowledgments}
This work has received funding from the European Union's Horizon Europe research and innovation programme under Grant Agreement No. 101168042 project TRIFFID (auTonomous Robotic aId For increasing First responders Efficiency) and No. 101189557 project TORNADO (foundaTion mOdels for Robots that haNdle smAll, soft and Deformable Objects).

\bibliographystyle{ieeetr}  % You can use plain, ieeetr, unsrt, etc.
\bibliography{references}    % This points to references.bib

\end{document}